\pdfoutput=1

\documentclass[11pt]{article}

\usepackage[final]{acl}

\usepackage{times}
\usepackage{latexsym}

\usepackage[T1]{fontenc}

\usepackage[utf8]{inputenc}

\usepackage{microtype}

\usepackage{inconsolata}

\usepackage{graphicx}

\usepackage{subcaption,booktabs}
\usepackage{tabularx}
\usepackage{multirow, amsmath}
\usepackage{placeins}
\usepackage{gensymb} 
\usepackage{amssymb}
\usepackage{amsthm}
\usepackage{color, colortbl}
\usepackage{float}
\usepackage{enumitem}
\usepackage{soul}
\usepackage{xcolor}
\usepackage{hyperref}
\usepackage{pgfplots}
\usepackage{tikz}
\usetikzlibrary{patterns}
\usepackage{setspace}
\usepackage{hyperref}
\usepackage{longtable}
\usepackage{supertabular}
\usepackage{xltabular}

\definecolor{bblue}{HTML}{4F81BD}
\definecolor{rred}{HTML}{C0504D}
\definecolor{ggreen}{HTML}{9BBB59}
\definecolor{ppurple}{HTML}{9F4C7C}

\newcommand*{\affmark}[1][*]{\textsuperscript{#1}}

\title{Will the Prince Get True Love's Kiss? \\ On the Model Sensitivity to Gender Perturbation over Fairytale Texts}

\author{
Christina Chance\affmark[\dag]
\quad Da Yin\affmark[\dag]
\quad Dakuo Wang\affmark[\ddag]
\quad Kai-Wei Chang\affmark[\dag]
\\
{\affmark[\dag]University of California, Los Angeles} \quad
{\affmark[\ddag]Northeastern University} \quad 
\\
\texttt{\{cchance, da.yin, kwchang\}@cs.ucla.edu}\\
\texttt{d.wang@northeastern.edu}\\
}

\begin{document}
\maketitle
\begin{abstract} 
In this paper, we study whether language models are affected by learned gender stereotypes during the comprehension of stories. Specifically, we investigate how models respond to gender stereotype perturbations through counterfactual data augmentation. Focusing on Question Answering (QA) tasks in fairytales, we modify the \texttt{FairytaleQA} dataset by swapping gendered character information and introducing counterfactual gender stereotypes during training. This allows us to assess model robustness and examine whether learned biases influence story comprehension. Our results show that models exhibit slight performance drops when faced with gender perturbations in the test set, indicating sensitivity to learned stereotypes. However, when fine-tuned on counterfactual training data, models become more robust to anti-stereotypical narratives. Additionally, we conduct a case study demonstrating how incorporating counterfactual anti-stereotype examples can improve inclusivity in downstream applications.

\end{abstract}

\section{Introduction} \label{introduction}

Fairytales, traditionally an oral form of storytelling, are used in various cultures as a way to pass cultural norms and practices down through generations. They are used in the classroom and at home to teach children reading comprehension, story structure, and develop cultural literacy~\cite{cinderella_classroom}.

However, it is known that there are strong gender biases within fairytales, specifically embodied through stereotypes. Included in the cultural norms defined in fairytales are gender roles and expectations; however, while cultural norms have evolved over time, these fairytales have not, yet are still purposed in the same way as decades earlier \cite{isaza2023fairy}. In many of these fairytales, the storyline consists of a brave, strong prince rescuing a distressed, helpless princess with true love's kiss. These stories are filled with gender biases that cast harmful and limiting stereotypes on various demographics. Studies have shown that repeatedly presenting children with gender bias stereotypes has a negative impact on their confidence and places limitations on their ability \cite{impact}. 

In NLP, fairytales are useful for assessing narrative comprehension of models due to the quantity and cultural diversity of fairytales as well as their well-studied use in education. Although existing question answering (QA) models perform well on fairytale datasets like \texttt{FairytaleQA} ~\cite{xu-etal-2022-fantastic}, we are curious how much these models rely on learned gender stereotypes and if these models will have consistent performance when presented with anti-gender stereotypes. Inspired by the literature on gender bias evaluation and mitigation~\cite{hall-maudslay-etal-2019-name}, we create counterfactual datasets to disrupt any learned gender-biased correlations from the pre-trained Language Models (LMs) and pre-trained Large LMs (LLMs). Specifically, we conduct our studies on \texttt{FairytaleQA}, a narrative comprehension dataset for children in kindergarten to eighth grade. 

\begin{figure}[t]
    \centering
\fbox{\includegraphics[width=0.95\linewidth]{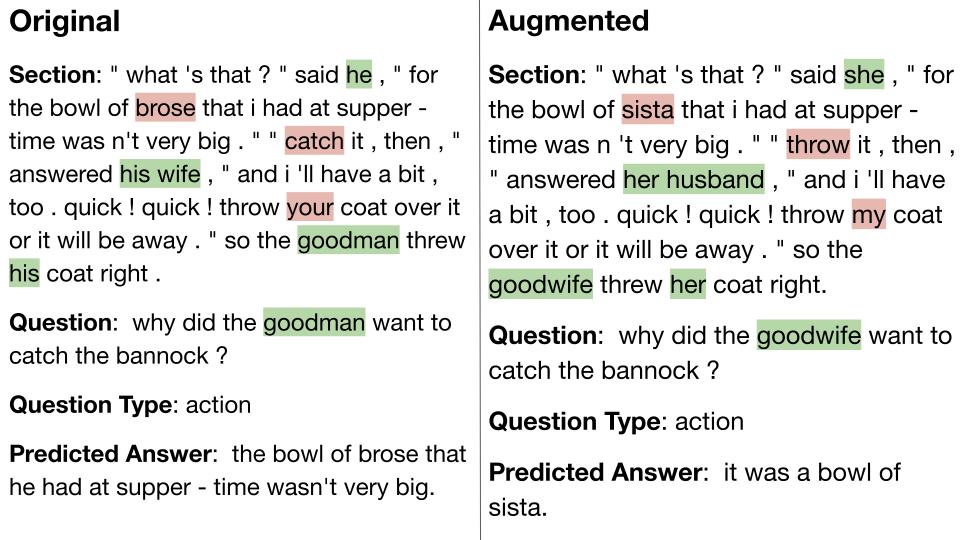}}
    \caption{Original and counterfactual test example using the LLM-assisted rule-based translation approach. The prediction of the \texttt{FairytaleQA}  model significantly changes after gender perturbation.}
    \label{fig:enter-label}
\end{figure}

To create the counterfactual dataset, we use three different approaches for data augmentation, including rule-based translation, LLM rewriting, and LLM-assisted rule-based translation. The last two approaches leverage the power of LLMs to support comprehensive rewriting for any text-domain. These approaches perform gender perturbations by swapping gendered nouns such as \emph{queen}: \emph{king} while preserving the ground truth labels. 

The evaluation experiment initially tests pre-trained LMs fine-tuned on the original \texttt{FairytaleQA} dataset. These models are then assessed on counterfactual test data synthesized with different approaches. The results reveal a consistent drop in performance, indicating a learned bias in the pre-trained LMs. Next, we assess pre-trained LMs fine-tuned on the counterfactual \texttt{FairytaleQA} dataset, testing them on the corresponding test data. Although there's a slight drop in the original \texttt{FairytaleQA} test set's performance, consistent improvements are observed across counterfactual test sets. This suggests that while the overall model accuracy may decrease marginally, the model is able to robustly handle changes in the character information. Furthermore, when these models are fine-tuned on a combined random 50\% original and 50\% counterfactual \texttt{FairytaleQA} dataset, they outperform models solely fine-tuned on the counterfactual dataset. This demonstrates that fine-tuning on both counterfactual and original data supports both normative and counterfactual gender roles. Additionally, we perform a small case study to highlight the benefits of incorporating anti-stereotype examples into datasets in the context of diverse fairytale generation. This study introduces an innovative approach to counterfactual data augmentation, emphasizing more generalizable methods of counterfactual data generation and the importance of including counterfactual examples within a dataset.

\section{Related Works} \label{related works}
\noindent
\textbf{Gender Biases in Fairytales.} Within education and the social sciences, work assessing the impact of gender bias in children's stories has shown a detrimental impact on children's self-esteem. \citet{cinderella_classroom} showed that both girls and boys age 9 to 11 benefited from stories that featured characters who shared their gender identity as a hero. This work has led to the re-framing of many stories within the classroom \cite{against_grain}, like \citet{diverge_traditional} who showed that introducing anti-sexist picture books to young girls led to them beginning to challenge the status quo and push back on the gender biases they were experiencing in their life, displaying the effect of these stories. 

\noindent
\textbf{Language Model Gender Bias.} Language models capture subtle and overt biases from the training corpus which propagates through the use of the model \cite{stochastic_parrots}. Many works have evaluated gender bias in contextualized word embeddings and have presented various approaches around removing gender associations in non-gendered words \cite{basta-etal-2019-evaluating,zhao-etal-2019-gender,bolukbasi2016man,cheng-etal-2022-debiasing}. Other methods have been suggested, such as debiasing co-reference resolution \cite{zhao-etal-2018-gender} or adversarial learning to debias dialogue generation \cite{liu-etal-2020-mitigating}, but no approaches have had success in complete debiasing \cite{gira-etal-2022-debiasing}. Although these approaches have reduced gender biases in LLMs, they address only occupational biases. Occupation bias refers to a form of discrimination based on a person's job or occupation, for example the assumption that all nurses are female and all doctors are male. We chose to focus on \textit{FairytaleQA} and fairytale texts because the gender biases found in these texts, including stereotypes and microaggressions, are out-of-domain from biases used in current bias mitigation approaches. 

\noindent
\textbf{Counterfactual Data Augmentation.} Proposed by \citet{lu2019gender}, Counterfactual Data Augmentation (CDA) is a corpus augmentation strategy that performs transformations on the data to break underlying gender-biased correlations in the model while preserving ground truth labels and accuracy. CDA is used in various works to address occupational bias \cite{hall-maudslay-etal-2019-name}. Many works aimed to mitigate occupational bias as these biases are more apparent in word embeddings. Other works have suggested improvements on the original CDA method including \cite{hall-maudslay-etal-2019-name} which proposed Counterfactual Data Substitution (CDS) which addresses duplication of text and name intervention. In \citet{qian-etal-2022-perturbation-v2}, a conditional seq2seq model is used to perform perturbations across various demographic axes, including gender, race/ethnicity, and age, both to assess model sensitivity and to develop the PANDA dataset. While their work provides a strong baseline, our approach differs in key ways: we adopt a different strategy for selecting which terms to perturb, leverage LLMs as the perturbation method, and focus on a specific domain, fairytales, to illustrate the broader implications of counterfactual data. Our goal in this work is to create a CDA approach that supports perturbations beyond those developed to address occupation bias, including stereotype biases and microaggressions in which the biases are more present in latent themes throughout the storylines.

\section{Approaches \& Evaluation} \label{approaches}
We utilize CDA to perform gender perturbations for the \texttt{FairytaleQA}  dataset. We perturb all pronouns and any gendered word, such as \emph{princess} or \emph{seamstress}, to its binary opposite gender, \emph{prince} and \emph{tailor}, respectively. These gender perturbations are applied to the story section, the question, and the answer. We assess various approaches to CDA to find the most precise and robust approach. The approaches presented vary within those two assessment qualities as rule-based translation allows for more controlled augmentations, and LLM rewriting allows for a more robust and expansive dictionary of gender pairs.   

The \texttt{FairytaleQA} dataset is a narrative comprehension resource designed for students from Kindergarten to eight grade. The datatset is a collection of 278 culturally diverse fairytales and with 10,580 questions. The question types are broken down to cover seven narrative elements -- setting, character, action, outcome resolution, feeling, causal relationship, prediction --  of a story \cite{xu-etal-2022-fantastic}. Examples of questions for each category type can be found in the Appendix in Table ~\ref{question_type_examples}. For more aggregated analysis due to size of the test set, we further classify the question types into abstractive and extractive question types. We define abstractive question types as those in which the answer is not explicitly in the text but requires the model to use the context provided in the section. Abstractive question types include outcome resolution, causal relationship, and prediction. Extractive question types are those in which the answer is explicitly given in the text. These question types include setting, character, action, and feeling. While this stratification partially aligns with the explicit and implicit labels in the original dataset, those are assigned on the question level while ours is assigned on the question types level, as we found that extractive questions were more susceptible to performance drops due to these perturbation compared to abstractive questions which would in cases benefit from the perturbations.

\noindent\textbf{Bias Scoring.} We use the bias score as a method to delineate performance disparities across the various datasets and methods. In our study, we argue that the biases inherent in fairytales are not only perpetuated but also intensified by the model's pre-existing biases. To assess the robustness of the models to lexical level perturbation, we verify that the model maintains consistency in output despite gender augmentations. 

\noindent\textbf{Consideration of Names During CDA.} We will not swap proper names during CDA approaches. We aim to break gendered associations with proper names. For instance, assigning he/him pronouns to a name like \emph{Cinderella} challenges the model's default female association with the name. Moreover, language models are primarily trained on Eurocentric and Western text, whereas the \texttt{FairytaleQA} dataset is culturally diverse. Assuming gender based on names from cultures or regions not included in the model's training data can introduce additional biases. While some cultures have naming conventions, there are often exceptions, and some names are gendered differently across cultures, making accurate predictions challenging without sufficient cultural context \cite{gautam-etal-2024-stop-v2}. Expecting the model to predict gender accurately across various cultures contradicts the purpose of our work. 

\subsection{Counterfactual Perturbation Methods}
\noindent\textbf{Rule-based translation.} The rule-based translation approach to developing the counterfactual dataset follows the approach in \citet{zhao-etal-2018-gender} which utilizes a dictionary of gendered word pairs and pronoun pairs, we additionally add gendered word pairs such as \emph{heir}: \emph{heiress}. We include both words in a pair as a keys, so \emph{heiress}: \emph{heir} would additionally be in the dictionary. We do not include proper nouns in the dictionary. We then iterate through each token in the data checking if the token is in the dictionary. If the token is in the dictionary as a key, we take the value of the key-value pair to replace the key in the text. We also include special checks for the pronoun \emph{her} in which if the token word is \textit{her} and the part of speech is personal pronoun, we swap with \emph{him}, otherwise with \emph{his}. The original dictionary was curated by Amazon MTurk workers, but the dictionary used in this work is modified to support gendered language associated with fairytales like the pair \emph{seamstress}: \emph{tailor}. 

This approach allows intentional control over what words are being modified as well as what they are being modified to. The limitation of this approach is that the translation is based on a static dictionary and therefore would need a new dictionary created for each new domain. 

\noindent\textbf{LLM text rewriting.} The LLM text rewriting approach uses the power and knowledge of LLMs like \verb|gpt-3.5-turbo| to perform gender word translations. We use a prompt, as shown in the Appendix in Table~\ref{story_rewrite_prompt}, to instruct generative LLMs to perform gender augmentation on the fairytale section, question, and answer while maintaining the original formatting. This approach allows for a larger scope for possible word augmentation. One limitation of this approach is, while the prompt specifies that only gendered nouns should be modified, the model also modifies gendered adjectives and lacks consistency in what words are modified. In the example below, we see that the LLM (i.e., \verb|gpt-3.5-turbo| in this work), modified adjectives pertaining to the non-gendered noun "couple". \\
\underline{Original}: "they were a very canty and contented couple, for they had enough to live on, and enough to do ." \\
\underline{LLM Perturbed}: "they were a very cheerful and contented couple, for they had enough to live on, and enough to do ."

\noindent\textbf{LLM-assisted rule-based translation.} The LLM-assisted rule-based translation approach combines the precision of rule-based translation with the adaptability of LLM text rewriting, overcoming some limitations of both methods. Initially, an empty dictionary is used. We utilize NLTK's part-of-speech (POS) tagger which classifies each word in the sentence as their associated POS. We do this for each section, question, and answer of a test case set. Then when the program encounters a noun or pronoun, it checks if the word is in the dictionary. If the word is in the dictionary, the standard rule-based translation is applied. If not, the word and a prompt (as described in the Appendix in Table~\ref{LM_assisted_prompt}) are processed through a LLM to generate the opposite binary gender word. This pair is then added to the dictionary, which is saved and reused in subsequent runs. The approach's drawback lies in the computational and financial costs, as well as the unpredictability and inconsistency of LLMs.

\noindent\textbf{CDA Human Evaluation.} To evaluate the quality of the perturbations done using a LLM, we perform an evaluation on subset of the data done by one of the authors. In this evaluation, we take a subset of the same 50 samples from the test data for two different CDA approaches. We evaluate this sample on four criteria on a scale of 1 (not at all) to 5 (always). The criteria assess the following:
\begin{itemize}[noitemsep,topsep=0pt]
    \item \textbf{Quality of swap.} Does the gendered word pair semantically and contextually make sense and are the correct words swapped?
    \item \textbf{Consistency of storyline.} Does the storyline remain the same and make sense despite the gender perturbations?
    \item \textbf{Consistency of swaps.} Does the approach perturb the same word with the same counterfactual pair every time?
    \item \textbf{Grammar.} Is the story grammatically correct?
\end{itemize}
Using this criterion, we evaluated the LLM-based rewriting approach utilizing \verb|gpt-3.5-turbo|. While content consistency and grammatical correctness received perfect average scores of 5.0/5.0, the quality of swaps averaged 4.08/5.0, and the consistency of swaps scored 4.76/5.0. Across the dataset, we observed inconsistent augmentations of the same text. For instance, the model frequently failed to perform standard swaps, such as \emph{father} to \emph{mother}. Additionally, it often inferred gender for neutral words like \emph{pink} and \emph{blue} or \emph{angel} and \emph{demon}. In some cases, the plot was altered to align with gender expectations—for example, rewriting a female character to avoid going to war, despite the original male character engaging in battle. In comparison, we assessed the LLM-assisted rule-based translation approach, also using \verb|gpt-3.5-turbo|, and observed improvements in swap quality (4.64/5.0) and swap consistency (4.86/5.0). However, grammatical accuracy dropped slightly to 4.5/5.0 due to errors in possessive pronoun perturbation. Additionally, the lack of contextual understanding in this approach limited the model’s ability to handle less common words, such as "brose," as shown in Figure~\ref{fig:enter-label}. In this instance, the NLTK tokenizer, used for preprocessing before querying the LLM, splits "didn't" into "didn" and "t." The LLM then misinterpreted "didn" as a gendered word and generated "dida" as its assumed opposite-gender counterpart. We selected NLTK because it allowed us to create a regex-based tokenizer capable of handling the special characters present in the dataset, which other tokenizers did not support.

\begin{table*}[ht!]
    \small
    \scalebox{0.67}{
    \begin{tabular}{lcccc|cccc|cccc|cccc}
    \toprule
    & \multicolumn{4}{c}{\textbf{Original Data}} & \multicolumn{4}{c}{\textbf{Augmented Data}} & \multicolumn{4}{c}{\textbf{\begin{tabular}[c]{@{}c@{}}50\% Original + \\ 50\% Augmented Data\end{tabular}}} & \multicolumn{4}{c}{\textbf{\begin{tabular}[c]{@{}c@{}}Full Original + \\ Full Augmented Data\end{tabular}}} 
     \\
     \cmidrule{2-5}\cmidrule{6-9}\cmidrule{10-13}\cmidrule{14-17}
    \textbf{\begin{tabular}[c]{@{}l@{}}Question \\ Type\end{tabular}} & \textbf{Orig.} & \textbf{\begin{tabular}[c]{@{}c@{}}Rule- \\ Based\end{tabular}} & \textbf{\begin{tabular}[c]{@{}c@{}}LLM\\ Rewrite\end{tabular}} & \textbf{\begin{tabular}[c]{@{}c@{}}LLM-\\ Assisted \\ Rule- \\ Based\end{tabular}} & \textbf{Orig.} & \textbf{\begin{tabular}[c]{@{}c@{}}Rule- \\ Based\end{tabular}} & \textbf{\begin{tabular}[c]{@{}c@{}}LLM\\ Rewrite\end{tabular}} & \textbf{\begin{tabular}[c]{@{}c@{}}LLM- \\ Assisted \\ Rule- \\ Based\end{tabular}} & \textbf{Orig.} & \textbf{\begin{tabular}[c]{@{}c@{}}Rule- \\ Based\end{tabular}} & \textbf{\begin{tabular}[c]{@{}c@{}}LLM\\ Rewrite\end{tabular}} & \textbf{\begin{tabular}[c]{@{}c@{}} LLM- \\ Assisted \\ Rule- \\ Based\end{tabular}} & \textbf{Orig.} & \textbf{\begin{tabular}[c]{@{}c@{}}Rule- \\ Based\end{tabular}} & \textbf{\begin{tabular}[c]{@{}c@{}}LLM\\ Rewrite\end{tabular}} & \textbf{\begin{tabular}[c]{@{}c@{}} LLM- \\ Assisted \\ Rule- \\ Based\end{tabular}}  \\
    
    \toprule
    
    ALL &           65.29  & 64.38   & 63.53   & 62.66 & 64.56 & \textbf{64.80} & 63.54 & \textbf{63.44} & 65.04 & \textbf{65.07} & \textbf{64.19} & \textbf{63.28} & 64.84 & \textbf{64.63} & 63.24 & \textbf{63.09} \cr
    \midrule
    Setting &       89.61  & 87.87    & 87.92    & 85.25 & 90.15 & 88.66 & 86.69 & 86.13 & 89.61 & 88.40 & 88.54 & 84.06 & 90.18 & 88.88 & 86.61 & 84.46 \cr
    Character &     85.67  & 83.86    & 79.77   & 82.10 & 84.12 & 85.18 & \textbf{83.09} & 82.98 & 85.02 & \textbf{87.24} & 81.95 & 83.75 & 84.65 & 84.41 & 80.16 & 84.07 \cr
    Action &        74.70  & 74.48    & 72.84   & 72.08 & 74.38 & 74.44 & 72.66 & \textbf{73.01} & 74.59 & 74.29 & \textbf{73.92} & \textbf{73.19} & \textbf{76.11} & \textbf{75.67} & \textbf{74.58} & \textbf{73.22} \cr
    Outcome Res. &  56.94  & 53.80    & 55.00    & 57.46 & 57.83 & 56.79 & 55.06 & 58.03 & 56.57 & 54.52 & 55.36 & 54.91 & 56.71 & 56.91 & 56.88 & 57.57 \cr
    Feeling &       49.41  & 48.47    & 47.49    & 43.75 & 48.74 & 46.28 & 45.37 & 43.45 & 48.48 & 47.51 & 46.52 & 44.68 & 50.05 & 47.21 & 45.61 & 44.38 \cr
    Causal Rel. &   56.98  & 56.40    & 56.64    & 55.29 & 54.53 & 56.53 & 56.44 & 55.19 & 55.90 & \textbf{57.57} & 56.67 & 55.50 & 53.42 & 55.65 & 54.68 & 54.77 \cr
    Prediction &    35.57  & 34.98    & 35.28    & 33.31 & 38.25 & 38.20 & 36.44 & 38.92 & 39.92 & 36.38 & 37.20 & 36.64 & 37.46 & 32.74 & 32.23 & 33.10\cr
    \bottomrule
    \end{tabular}
    }
    \caption{\label{t5-rouge-full} \small ROUGE-L F1 scores for the T5 model fine-tuned on the ruled-based gender augmented \texttt{FairytaleQA}  dataset (center l.h.s.) and 50\% of original + 50\% of rule-based gender augmented \texttt{FairytaleQA}  dataset (center r.h.s.), and full original + full rule-based gender augmented \texttt{FairytaleQA}  dataset (r.h.s.) and tested on the rule-based gender augmentation, LLM gender augmentation, and LLM assisted rule based gender augmentation test dataset. Bold values indicate a statistically significant increase to the 95\% confidence compared to ROUGE-L F1 scores for the T5 model fine-tuned on the original dataset (l.h.s.).}
    
\end{table*}

\section{Experiments} \label{experiments}

We used two pre-trained LMs as our base models for fine-tuning on \texttt{FairytaleQA}: T5 \cite{raffel2020exploring} and BART \cite{lewis-etal-2020-bart}. For both models, we set the number of epochs to 4, the learning rate to $5 \times 10^{-5}$, and the seed to 88 for reproducibility. We chose our hyperparameters and LMs based on the prior work done by \citet{xu-etal-2022-fantastic} in order to attempt to reproduce those results as our base for testing. Additionally, we chose to assess only T5 and BART to test sensitivity based on fine-tuning. This approach is necessary for our task, as these stories are culturally robust and therefore out of domain for even the best LLMs, as suggested in the discussion of name considerations for CDA. 

In the original work, they chose BART as the backbone of their fine-tuned models since it had the best performance for the QA task using \texttt{FairytaleQA}. In our own analysis, we found that finetuned T5 actually outperformed finetuned BART for the \texttt{FairytaleQA} task, so we additionally used T5. We used the ChatGPT API (\verb|gpt-3.5-turbo|) as our LLM model for augmentation since the ChatGPT API system is built on the \texttt{GPT 3.5 turbo} architecture \cite{openai_chatgpt}. We assess these models on the original and counterfactual \texttt{FairytaleQA} and report the ROUGE-L \cite{lin-2004-rouge} F1 scores as the model performance and use a pooled t-test to compare the results of varying fine-tuned models. Additionally, we had a training/validation/testing split of 8:1:1 that was taken from the original dataset, in which the testing data is unseen by the model at test time.

\subsection{Assessing Model Sensitivity to Gender Perturbations}
To assess the sensitivity of the models fine-tuned on the \texttt{FairytaleQA} dataset, we perform three augmentation methods on the training, validation, and test datasets in which we swap gendered nouns and pronouns, such as occupations or familial titles, to the opposite gender and do not modify proper names such as \emph{Cinderella}. To perform this augmentation, we use the approaches mentioned in Section \ref{approaches}. We run two sets of experiments to assess stereotype bias in the dataset. The first experiment assesses the sensitivity of the model fine-tuned with original data and the second experiment assesses the sensitivity of the model fine-tuned with counterfactual data.

The test set size is limited due to challenges in recruiting additional educational experts and obtaining more examples. While we highlight statistically significant values in several tables, the small sample size makes statistical significance harder to achieve overall. As a result, we focus our discussion on performance trends based on changes in accuracy.

\noindent\textbf{Gender Bias in Fairytale QA Models.} 
We assess T5 and BART fine-tuned on the original \texttt{FairytaleQA} training and validation data and test on the original and counterfactual test data using each approach to assess the baseline biases in the dataset. In Table~\ref{t5-rouge-full}, we see across all augmented test sets a drop in the performance of the T5 model compared to the original test data suggesting that the model possesses some learned biases when fine-tuned on the original \texttt{FairytaleQA} dataset. We additionally witness fairly consistent drops across all question types as well, with some question types such as character and action having more pronounced performance drops. For BART, as seen in Table~\ref{bart-rouge-full} in the Appendix, we also witness consistent changes in the performance of the model with some counterfactual data outperforming the original data. The table suggest that the small perturbation of changing the gender of characters in the testing set has an impact on models' performance for the task of question answering.

\noindent\textbf{Fine-Tune with Gender CDA.} 
We fine-tune T5 and BART using three different combinations of the original and counterfactual \texttt{FairytaleQA} datasets to observe the impact of integrating counterfactual training and validation into the original sets. In the first combination, we use the complete counterfactual training and validation sets. The second consists of random 50\% of the original and 50\% of the counterfactual training and validation sets. We avoid the duplication of questions and maintain the distribution of question types in the dataset. The third is the full original and full counterfactual training and validation \texttt{FairytaleQA} combined. We evaluate the models using the same test data as the previous experiment for all three training and validation combinations. We opt for the last two sets for fine-tuning to prevent potential overfitting caused by doubling the training size when combining the full and counterfactual sets. Additionally, we compare the same test set against different fine-tuning training sets to avoid comparing different test set performance. 

In Table~\ref{t5-rouge-full}, we compare the T5 model's performance after fine-tuning on either the counterfactual or the original training/validation set. The results indicate improved performance across all counterfactual test sets. Notably, the model fine-tuned on the random 50\% counterfactual and 50\% original training/validation set (depicted in Figure~\ref{fig:counterfactual_original_bar}) shows a smaller performance drop for the original test set. BART shared similar performance differences as the T5 model as shown in the Appendix in Table~\ref{bart-rouge-full} and Figure~\ref{fig:counterfactual_bar_BART}.


\subsection{Inclusive Fairytale Generation: Case Study}
To further investigate the impact of counterfactual data augmentation, we conduct a case study examining how incorporating anti-stereotype examples into datasets influence fairytale generation. While previous sections focus on evaluating the robustness of fairytale comprehension models, this section extends the discussion by exploring how counterfactual augmentation can actively shape the narratives produced by generative models. By doing som we aim to assess whether introducing counterfactual data mitigates bias reinforcement and fosters more inclusive story generation. 

In particular, we examine a scenario where gender biases present in training data contribute to further bias propagation within generated fairytales. Fairytale generation is a crucial example for this analysis because stories play a formative role in shaping children's perceptions of gender roles. If a model trained on traditional fairytales associates heroism with male characters and passivity with female characters, it risks perpetuating these biases in newly generated narratives. By augmenting training data with counterfactual gender examples, we can assess whether such approaches lead to more balanced and diverse representations.

Beyond bias mitigation, counterfactual augmentation aligns with broader goals in creative AI and education. Large language models (LLMs) are increasingly used to generate children's stories, and ensuring diverse representation within these narratives is essential. Counterfactual gender augmentation allows young readers to encounter protagonists of all genders as adventurers, leaders, scientists, and superheroes, challenging traditional norms. Moreover, leveraging LLMs for inclusive story generation enhances scalability and cultural adaptability, making it possible to generate narratives that better reflect the diverse experiences of students.

This aligns with the principles of culturally responsive teaching, which emphasizes the importance of culturally relevant content in education \cite{gay2018culturally}. Studies suggest that students engage more deeply with narratives they can relate to, particularly when these stories reflect their identities and lived experiences. By equipping generative models with anti-stereotype training data, we can produce fairytales that are not only more inclusive but also more meaningful for diverse student populations.

To evaluate this, we prompt \verb|gpt-3.5-turbo| to generate new fairytales inspired by either the counterfactual or original stories from the test dataset. The model's system  role is instructed as follows: "You are a creative writer for children's stories. Given the current story, write a new story while maintaining the lessons and beliefs." The model's user role is instructed as follows: "Current story: *Insert Story Section* Write a new children's fairytale inspired by the current story." The max generation length was 700 tokens and temperature was set to 0.7. For assessment, we use the following set of metrics:
\begin{itemize}[noitemsep,topsep=0pt]
    \item \textbf{Repetitive Plot}: Repeats similar text, sentence structures, or adjectives.
    \item \textbf{Unrelated Events}: Introduces unrelated characters or actions; omits key characters or scenes.
    \item \textbf{Conflicting Logic}: Contains incorrect temporal relationships or contradictions.
    \item \textbf{Poor Continuity}: Difficult to follow due to inconsistencies.
    \item \textbf{Unsafe Content}: Includes material that may be inappropriate for children.
    \item \textbf{Bias Propagation}: Reinforces stereotypes or gender role expectations.
\end{itemize}

We assess narrative quality at both the local and global levels, drawing inspiration from \cite{guan-etal-2021-openmeva-v2} to evaluate continuity, clarity, logic, and coherence. Additionally, we analyze the generated text for safety \cite{ermolaeva-etal-2024-tame-v2} and inclusive language. Our overall evaluation score starts at 6, with point deductions ranging from 0 to -2 per metric, depending on severity.

Using these metrics, we consider good stories as those that maintain a similar lesson learned from the provided story while still creating a new storyline with characters, a new adventure or challenge faced, and a character arc. The goal is for the stories to still be exciting and interesting. Due to the comprehension level, we also want to ensure consistency and continuity in the storyline and therefore penalize for holes or leaps in the plot as well as unnecessary and unrelated information which may cause confusion. We provide examples of generated stories that are penalized for each metric in the Appendix in Table~\ref{generated_stories_example}.

A single annotator analyzed 30 pairs of generated stories, each prompted with either the original or counterfactual story section. On average, stories generated from the original section received a score of 4.933/6.0, while those generated from the counterfactual section scored 5.67/6.0. Table~\ref{case-study} presents the average point deductions for each metric across both conditions. The average difference between original and counterfactual generations, calculated by subtracting the counterfactual score from the original score, was -0.733.

Beyond numerical evaluation, qualitative observations highlighted notable trends. The annotator noted limited diversity in character names (e.g., frequent use of Lily, Fin, Luna, and Pip) and the overall structural similarity across stories, often exhibiting minimal narrative development. In story pairs with similar plots, descriptions in the original generations frequently emphasized physical appearance, whereas counterfactual generations leaned toward personality traits. Additionally, stories prompted by the original section exhibited more logical inconsistencies, with abrupt scene transitions that assumed unstated details. These generations also adhered more rigidly to the style and conventions of traditional fairytales, occasionally misusing adjectives that, while contextually incorrect, were commonly associated with specific characters, animals, or roles in classical storytelling. This pattern was significantly less prevalent in stories generated from the counterfactual section.

\begin{table}[ht!]
    \centering
    \footnotesize
    \begin{tabular}{lcc}
        \toprule
        \textbf{Metric}            & \textbf{Original} & \textbf{Counterfactual} \\
        \midrule
        Repetitive Plot   & 0.067             & 0.033                   \\
        Unrelated Events  & 0.233             & 0.167                   \\
        Conflicting Logic & 0.467             & 0.133                   \\
        Poor Continuity   & 0.133             & 0                       \\
        Unsafe Content    & 0                 & 0                       \\
        Bias Propagation  & 0.167             & 0                       \\
        \bottomrule
    \end{tabular}
    \caption{\label{case-study} The average point deduction per metric based on the prompted story section.}
\end{table}

\section{Discussion} \label{discussion}

\noindent\textbf{Question Type Analysis.} 
We stratify the test set by question type for further analysis. Table~\ref{question-distribution} presents a breakdown of the number of test cases per category that showed a statistically significant change in performance, as determined by the pooled T-test. This comparison evaluates the original test set against various CDA approaches for the T5 model fine-tuned on the original FairytaleQA dataset. We use this stratification as well as the discussion of abstractive and extractive question types to further discuss the results. While we see in Table~\ref{t5-rouge-full} an overall drop across both classes of questions for T5, we witness that abstractive questions have a smaller performance drop compared to extractive questions. A possible cause is that the model has to reason beyond learned bias correlation in order to successfully predict the answer. Due to the anti-stereotype perturbations in the test set, the model is not able to rely of the learned correlations on gender because they were not present in the fairytale sections provided so it has to develop a new understanding based on the text. However, for the extractive questions, the anti-stereotype question answer pairs are more difficult to accurately predict because it is not logically supported by the current model understanding. This drop in extractive questions shows a reliance on prior learned gender correlations as these questions are information extraction, a task that models tend to have state of the art performance for. 

\noindent\textbf{LLM Output Quantitative Assessment.} Using the outputs of the models fine-tuned on the original \texttt{FairytaleQA} dataset, we compare the generated output of the original test data with that of the gender-perturbed test data to see what type of changes are present in the predicted answer. We first collect the samples in which the ROUGE-L F1 score has a different accuracy compared to its unperturbed pair. We then use the BERTscore \cite{zhang2020bertscore} to assess the semantic similarity of the generated answer from the original and counterfactual data as we found that due to the gender perturbations, the ROUGE-L score does not successfully capture similarity in output. Using the BERTscore, we expect that the comparative BERTscore for the predicted answers in which only the gender differs have an about 1.0 BERTscore. With this understanding, we set a threshold of 0.5 for the BERTscore and flag the examples whose comparative BERTscores are lower than 0.5.   
Within this set of test examples, we found that there were very few instances of explicit bias produced by the model. For some examples, the model produced unrelated text for some counterfactual test examples. Examples of this behavior are in Table~\ref{quant_table}. In other instances, the model produced more detailed answers that better aligned with the ground truth answers.

\begin{table}[ht!]
\centering
\footnotesize
\begin{tabular}{lccc|c}
\toprule
\textbf{\begin{tabular}[c]{@{}c@{}}Question\\ Type\end{tabular}} & \textbf{\begin{tabular}[c]{@{}c@{}}Rule-\\ Based\end{tabular}} & \textbf{\begin{tabular}[c]{@{}c@{}}LLM\\ Rewrite\end{tabular}} & \textbf{\begin{tabular}[c]{@{}c@{}}LLM- \\ Assisted \\ Rule-\\ Based\end{tabular}} & \textbf{\begin{tabular}[c]{@{}c@{}}Total\\ Count\end{tabular}}  \\

\midrule

ALL & 99 & 119 & 132  & 1007 \\
\midrule
Setting  & 2 & 2 & 4 & 62 \\
Character  & 3 & 11 & 5 & 103\\
Action  & 22 & 27 & 33 & 315\\
Outcome Res. & 10 & 10 & 11 & 78\\
Feeling  & 0 & 0 & 1 & 106\\
Causal Rel. & 54 & 54 & 64 & 278 \\
Prediction & 8 & 15 & 14 & 65\\
\bottomrule
\end{tabular}
\caption{\label{question-distribution}\small Question type count for outputs flagged as significantly modified based on quantitative evaluation approach and the total number of questions per type in test set.}
\end{table}

\noindent\textbf{Evaluation of Augmented Gendered Adjectives by LLM.}
To provide a transparent quality check for the different CDA approaches used, we found word error rates that align with the number of incorrectly modified adjectives and nouns in both the LLM rewriting and LLM-assisted rule-based translation approaches. Roughly for LLM rewriting the count of modified adjectives and nouns is 2872 and the word error rate compared to rule-based translation as the ground truth is 0.0630 and match error rate is 0.0618. Roughly for LLM-assisted rule-based translation, the count of modified adjectives and nouns is 2638 and the word error rate compared to rule-based translation as the ground truth is 0.0537 and match error rate is 0.0537. These counts and metrics are for the testing dataset which has 1007 examples. The word error rate and match error rate account for all possible changes made including the adjectives that were not gendered nouns and pronouns. These error rates are relatively small. In further iterations, we plan to address the error rates for the LLM-assisted rule-based translation approach.

\begin{figure}[ht!]
\scalebox{0.9}{
\begin{tikzpicture}
    \begin{axis}[
        width  = 0.5*\textwidth,
        height = 7.5cm,
        major x tick style = transparent,
        ybar=1.2*\pgflinewidth,
        bar width=4pt,
        ymajorgrids = true,
        ylabel = {Performance Difference},
        symbolic x coords={All,Setting,Character,Action,Outcome Res.,Feeling,Causal Rel.,Prediction},
        xtick = data,
        x tick label style={rotate=30,anchor=east},
        ymin=-3,
        ymax=6,
        x tick label style={/pgf/number format/1000 sep=},
        legend cell align=left,
        y label style={at={(axis description cs:0.1,.5)},anchor=south},
        ticklabel style = {font=\small},
        legend style={
                legend columns =1,
                at={(0,1)},
                anchor=north west,
                column sep=1ex,
                fill=none,
                draw=none,
                nodes={scale=0.8, transform shape}
        }
    ]
        \addplot[style={bblue,fill=bblue,mark=none}]
            coordinates {(All,-0.25) (Setting,0.00) (Character,-0.65) (Action,-0.11) (Outcome Res.,-0.37) (Feeling,-0.93) (Causal Rel.,-1.08) (Prediction,4.35)};

        \addplot[style={rred,fill=rred,mark=none}]
             coordinates {(All,0.69) (Setting,0.53) (Character,3.38) (Action,-0.19) (Outcome Res.,0.72) (Feeling,-0.96) (Causal Rel.,1.17) (Prediction,1.40)};

        \addplot[style={ggreen,fill=ggreen,mark=none}]
             coordinates {(All,0.66) (Setting,0.62) (Character,2.18) (Action,1.08) (Outcome Res.,0.36) (Feeling,-0.97) (Causal Rel.,0.03) (Prediction,1.92)};

        \addplot[style={ppurple,fill=ppurple,mark=none}]
             coordinates {(All,0.62) (Setting,-1.19) (Character,1.65) (Action,1.11) (Outcome Res.,-2.55) (Feeling,0.93) (Causal Rel.,0.21) (Prediction,3.33)};

        \legend{Original,Rule-Based,LLM,LLM-Assisted Rule-Based}
    \end{axis}
\end{tikzpicture}
}
 \caption{\label{fig:counterfactual_original_bar}\small Performance difference of ROUGE-L F1 scores between T5 model fine-tuned on 50\% original + 50\% counterfactual \texttt{FairytaleQA}  and T5 model fine-tuned on original \texttt{FairytaleQA} , a positive value showing an increase in performance. Each colored bar represents the test set augmented with the given approach.}
\end{figure}

\section{Conclusion} \label{conclusion}

In this work, we evaluate the story comprehension of language models when exposed to counterfactual gender stereotypes using the counterfactual \texttt{FairytaleQA} dataset, generated through multiple CDA approaches (e.g., GPT-3.5 Turbo and rule-based methods). As shown in Table~\ref{t5-rouge-full}, our results indicate that models are sensitive to gender-perturbed data. However, fine-tuning on a combination of original and counterfactual data improves performance, demonstrating the benefits of counterfactual augmentation. We argue that incorporating counterfactual data is a beneficial practice with potential advantages for downstream tasks, and we introduce a novel CDA approach that is both generalizable and adaptable across diverse domains.

Additionally, we conduct a case study examining the quality and inclusivity of generated fairytales when prompted with gender-stereotypical versus counterfactual gender-stereotypical stories from the dataset. Our quantitative and qualitative analysis shows that fairytales generated from counterfactual prompts exhibit greater readability, improved continuity, and a stronger emphasis on character traits over physical attributes. Given the performance shifts observed in our experiments, our findings, supported by this case study, suggest that integrating counterfactual anti-stereotype examples is an effective strategy for mitigating bias and fostering inclusivity in downstream applications.


\section*{Limitations}
We acknowledge that our work operates within the normative gender binary, which excludes other marginalized groups, such as non-binary and gender non-conforming individuals. In future work, we plan to evaluate language models on a more inclusive set of gender biases.

Additionally, our analysis takes a singular approach to gender bias. However, fairytales contain more complex and intersectional forms of gender bias that can negatively impact young children, particularly those related to beauty standards. Cultural expectations of beauty intersect with other biases, including fatphobia, ageism, and colorism. Additionally present in fairytales are themes of elitism and classism, all having significant impact on the framing of class and wealth for children \cite{91f82b87-e85c-3c8f-b535-6fa06a6d3656}. Attempting to study one dimension of gender bias is a disservice \cite{lalor-etal-2022-benchmarking} and in future works, we plan to address many other dimensions with the understanding that many of these dimensions intersect other "-isms" and require more than small perturbations \cite{03b7b870-dd21-39ad-8e1b-1ec247813aea}.

\section*{Ethics Statement}
The goal of this work is to assess the gender biases learned in a model in the context of fairytale text. The work aims to bring light to the impact and usefulness of counterfactual data augmentation in helping develop more inclusive and anti-stereotype datasets. While our work is centered on the normative view of gender, it can hopefully provide framing to assess these biases outside of the standard gender binary setting. 


\bibliography{anthology,custom}

\appendix
\onecolumn
\newpage

\section{Additional Results and Figures}

\begin{figure}[h]
\centering
\scalebox{0.9}{
\begin{tikzpicture}
            \begin{axis}[
               width  = 0.5*\textwidth,
                height = 7.5cm,
                major x tick style = transparent,
                ybar=1.2*\pgflinewidth,
                bar width=4pt,
                ymajorgrids = true,
                ylabel = {Performance Difference},
                symbolic x coords={All,Setting,Character,Action,Outcome Res.,Feeling,Causal Rel.,Prediction},
                xtick = data,
                x tick label style={rotate=45,anchor=east},
                ymin=-3,
                ymax=6,
                x tick label style={/pgf/number format/1000 sep=},
                legend cell align=left,
                y label style={at={(axis description cs:0.1,.5)},anchor=south},
                ticklabel style = {font=\small},
                legend style={
                        legend columns =1,
                        at={(0,1)},
                        anchor=north west,
                        column sep=1ex,
                        fill=none,
                        draw=none,
                        nodes={scale=0.8, transform shape}
                }
            ]
                \addplot[style={bblue,fill=bblue,mark=none}]
                    coordinates {(All,-0.73) (Setting,0.54) (Character,-1.55) (Action,-0.32) (Outcome Res.,0.89) (Feeling,-0.67) (Causal Rel.,-2.45) (Prediction,2.68)};
        
                \addplot[style={rred,fill=rred,mark=none}]
                     coordinates {(All,0.42) (Setting,0.79) (Character,1.32) (Action,-0.04) (Outcome Res.,2.99) (Feeling,-2.19) (Causal Rel.,0.13) (Prediction,3.22)};
        
                \addplot[style={ggreen,fill=ggreen,mark=none}]
                     coordinates {(All,0.01) (Setting,-1.23) (Character,3.32) (Action,-0.18) (Outcome Res.,0.06) (Feeling,-2.12) (Causal Rel.,-0.20) (Prediction,1.16)};
        
                \addplot[style={ppurple,fill=ppurple,mark=none}]
                     coordinates {(All,0.78) (Setting,0.88) (Character,0.88) (Action,0.93) (Outcome Res.,0.57) (Feeling,-0.30) (Causal Rel.,-0.10) (Prediction,5.61)};
        
                \legend{Original,Rule-Based,LLM,LLM Assisted Rule-Based}
            \end{axis}
        \end{tikzpicture}
}
 \caption{\label{fig:counterfactual_bar_BART}\small Performance difference of ROUGE-L F1 scores between BART model fine-tuned on counterfactual \texttt{FairytaleQA}  and BART model fine-tuned on original \texttt{FairytaleQA} , a positive value showing an increase in performance. Each colored bar represents the test set augmented with the given approach.}
\end{figure}

\begin{table*}[ht!]
\small
    \scalebox{0.67}{
    \begin{tabular}{lcccc|cccc|cccc|cccc}
    \toprule
    & \multicolumn{4}{c}{\textbf{Original Data}} & \multicolumn{4}{c}{\textbf{Augmented Data}} & \multicolumn{4}{c}{\textbf{\begin{tabular}[c]{@{}c@{}}50\% Original + \\ 50\% Augmented Data\end{tabular}}} & \multicolumn{4}{c}{\textbf{\begin{tabular}[c]{@{}c@{}}Full Original + \\ Full Augmented Data\end{tabular}}} 
     \\
     \cmidrule{2-5}\cmidrule{6-9}\cmidrule{10-13}\cmidrule{14-17}
    \textbf{\begin{tabular}[c]{@{}l@{}}Question \\ Type\end{tabular}} & \textbf{Orig.} & \textbf{\begin{tabular}[c]{@{}c@{}}Rule- \\ Based\end{tabular}} & \textbf{\begin{tabular}[c]{@{}c@{}}LLM\\ Rewrite\end{tabular}} & \textbf{\begin{tabular}[c]{@{}c@{}}LLM-\\ Assisted \\ Rule- \\ Based\end{tabular}} & \textbf{Orig.} & \textbf{\begin{tabular}[c]{@{}c@{}}Rule- \\ Based\end{tabular}} & \textbf{\begin{tabular}[c]{@{}c@{}}LLM\\ Rewrite\end{tabular}} & \textbf{\begin{tabular}[c]{@{}c@{}}LLM- \\ Assisted \\ Rule- \\ Based\end{tabular}} & \textbf{Orig.} & \textbf{\begin{tabular}[c]{@{}c@{}}Rule- \\ Based\end{tabular}} & \textbf{\begin{tabular}[c]{@{}c@{}}LLM\\ Rewrite\end{tabular}} & \textbf{\begin{tabular}[c]{@{}c@{}} LLM- \\ Assisted \\ Rule- \\ Based\end{tabular}} & \textbf{Orig.} & \textbf{\begin{tabular}[c]{@{}c@{}}Rule- \\ Based\end{tabular}} & \textbf{\begin{tabular}[c]{@{}c@{}}LLM\\ Rewrite\end{tabular}} & \textbf{\begin{tabular}[c]{@{}c@{}} LLM- \\ Assisted \\ Rule- \\ Based\end{tabular}}  \\
    
    \toprule
    
    ALL & 59.12 & 59.73 & 59.02 & 59.31 & \textbf{60.28} & \textbf{59.97} & 58.70 & 59.31 & \textbf{60.40} & 58.97 & 58.88 & \textbf{59.65} & \textbf{60.59} & \textbf{60.68} & 59.20 & \textbf{60.07} \cr
    \midrule
    Setting & 85.49 & 84.30 & 86.60 & 81.83 & 81.85 & 83.55 & 87.03 & 81.83 & 81.67 & 83.80 & 85.12 & 81.83 & 81.53 & 81.21 & 82.85 & 80.64 \cr
    Character & 80.58 & 80.42 & 79.01 & 78.72 & 79.19 & 79.90 & 76.35 & 78.72 & 79.56 & 78.81 & 75.47 & 80.24 & 79.34 & 79.65 & 77.65 & 80.54 \cr
    Action & 65.17 & 66.68 & 65.91 & 63.96 & \textbf{67.19} & 66.96 & 66.08 & 63.96 & \textbf{68.53} & 65.95 & 66.42 & 64.70 & \textbf{68.05} & 67.41 & 66.23 & 66.33 \cr
    Outcome Res. & 57.07 & 60.69 & 57.79 & 53.44 & 56.89 & 56.52 & 53.91 & 53.44 & 55.09 & 52.24 & 58.08 & 58.67 & 59.08 & 59.26 & 61.69 & 57.09 \cr
    Feeling & 41.87 & 41.87 & 39.96 & 48.14 & 47.20 & 45.18 & 41.85 & 48.14 & 42.91 & 41.97 & 40.99 & 45.74 & 41.76 & 45.33 & 40.59 & 46.28 \cr
    Causal Rel. & 51.77 & 51.24 & 50.86 & 52.57 & 51.53 & 51.22 & 50.55 & 52.57 & 52.73 & 51.63 & 50.94 & 51.73 & \textbf{53.47} & \textbf{53.56} & 51.51 & 52.63 \cr
    Prediction & 32.74 & 34.20 & 35.11 & 38.69 & 39.14 & 37.63 & 36.10 & 38.69 & 38.06 & 37.21 & 35.13 & 39.16 & 37.76 & 35.62 & 33.66 & 35.54 \cr
    \bottomrule
    \end{tabular}
    }
    \caption{\label{bart-rouge-full}\small ROUGE-L F1 scores for the BART model fine-tuned on the ruled-based gender augmented \texttt{FairytaleQA}  dataset (center l.h.s.), 50\% of original + 50\% of rule-based gender augmented \texttt{FairytaleQA}  dataset (center r.h.s.), and full original + full rule-based gender augmented \texttt{FairytaleQA}  dataset (r.h.s.) and tested on the rule-based gender augmentation, LLM gender augmentation, and LLM assisted rule based gender augmentation test dataset. Bold values indicate a statistically significant increase to the 95\% confidence compared to ROUGE-L F1 scores for the BART model fine-tuned on the original dataset (l.h.s.).}

\end{table*}

\section{Prompts and Examples}

\begin{table*}[!b]
    \small
    \begin{tabularx}{\textwidth}{X}
    \toprule
    Providing a pair of words, if the first word is gendered, the paired word is the opposite gender of the original word, if the first word is gender neutral, the paired word is the same word as the first word. Provide the word to complete the pair.\\

    prince $\rightarrow$ princess\\
    daughter $\rightarrow$ son\\
    bus $\rightarrow$ bus\\
    person $\rightarrow$ person\\
    teacher $\rightarrow$ teacher\\

    \bottomrule
    \caption{\label{LM_assisted_prompt} Prompt for the LLM-assisted rule-based translation approach. Each noun in the fairytale, question, and answer set is passed and the model produces the word pair based on the input. }
        
    \end{tabularx}
    
\end{table*}
\begin{table*}[]
\small
\begin{tabularx}{\textwidth}{lX}
\toprule
\textbf{Question Type} & \textbf{Examples} \cr
\toprule
Setting &
  \begin{tabular}[c]{@{}l@{}}Where did the princess travel till?\\ Where did the princess fall asleep?\\ Where did the creature and the huntsmen go?\end{tabular} \cr
      \midrule

Character &
  \begin{tabular}[c]{@{}l@{}}What kind of hair did the wife have?\\ How did the king's daughter look? \\ Who did the king's wife send for when she felt that she would soon die?\end{tabular} \cr
      \midrule

Action &
  \begin{tabular}[c]{@{}l@{}}What did the king's daughter say she must have? \\ What did the cloak have to made of? \\ What did the king command the huntsmen to do?\end{tabular} \cr
      \midrule

Outcome Resolution &
  \begin{tabular}[c]{@{}l@{}}What happened when the king's wife fell ill?\\  What happened after the huntsmen seized the maiden?\\  What happened when the dance ended?\end{tabular} \cr
      \midrule

Feeling &
  \begin{tabular}[c]{@{}l@{}}How did the king's daughter feel when she heard his decision?\\ How did the princess feel when she awoke?\\ How did the cook feel when the king called him in?\end{tabular} \cr
      \midrule

Causal Relationship &
  \begin{tabular}[c]{@{}l@{}}Why couldn't one find anyone like the king's wife?\\  Why were the messengers sent far and wide?\\  Why did the councillors say the king had to marry again?\end{tabular} \cr
      \midrule

Prediction &
  \begin{tabular}[c]{@{}l@{}}What will happen if the Many-furred creature drops a hair in the soup?\\  What will the princess do in half-an-hour?\\ What will the king do with the beautiful maiden when he sees her again?
  \end{tabular}\\
 \bottomrule

\end{tabularx}
\caption{\label{question_type_examples} Examples of the various question type categories from the \texttt{FairytaleQA} test set. }
\end{table*}
\begin{table*}[!t]
\footnotesize
    \centering
    \scalebox{0.95}{
    \begin{tabular}{llllll}
    \toprule
    \textbf{Approach} &
      \textbf{Question} &
      \textbf{\begin{tabular}[c]{@{}l@{}}Ground Truth \\ Answer\end{tabular}} &
      \textbf{\begin{tabular}[c]{@{}l@{}}Question\\ Type\end{tabular}} &
      \textbf{\begin{tabular}[c]{@{}l@{}}Original\\ Prediction\end{tabular}} &
      \textbf{\begin{tabular}[c]{@{}l@{}}Counterfactual \\ Prediction\end{tabular}} \\
    
      \midrule
    \begin{tabular}[c]{@{}l@{}}LLM\\ Rewrite\end{tabular} &
      \begin{tabular}[c]{@{}l@{}}why didn't the \\ widowed mistress \\ tease the old \\ woman anymore ?\end{tabular} &
      \begin{tabular}[c]{@{}l@{}}the widowed \\ mistress became \\ angry at her .\end{tabular} &
      \begin{tabular}[c]{@{}l@{}}causal \\ relationship\end{tabular} &
      she was mad. &
      \begin{tabular}[c]{@{}l@{}}he was not \\ a beauty.\end{tabular} \\
    
    \midrule
      
    \begin{tabular}[c]{@{}l@{}}LLM\\ Rewrite\end{tabular} &
      \begin{tabular}[c]{@{}l@{}}why did maie want \\ to possess a \\ fairy cow ?\end{tabular} &
      \begin{tabular}[c]{@{}l@{}}she heard about \\ the fairy cow \\ from the young men .\end{tabular} &
      \begin{tabular}[c]{@{}l@{}}causal\\ relationship\end{tabular} &
      \begin{tabular}[c]{@{}l@{}}ahti was a \\ mighty king.\end{tabular} &
      \begin{tabular}[c]{@{}l@{}}he wanted to \\ be a princess.\end{tabular} \\
    
    \midrule
      
    \begin{tabular}[c]{@{}l@{}}LLM\\ Rewrite\end{tabular} &
      \begin{tabular}[c]{@{}l@{}} who was \\ the king's son ? \end{tabular}&
      the gentleman . &
      character &
      he. &
      \begin{tabular}[c]{@{}l@{}}the young \\ woman.\end{tabular} \\
    
    \midrule

    Rule-Based &
      \begin{tabular}[c]{@{}l@{}}what happened after \\ the king returned \\ to the cottage ?\end{tabular} &
      \begin{tabular}[c]{@{}l@{}}it was empty , and \\ none could inform \\ his where his bride \\ had gone .\end{tabular} &
      \begin{tabular}[c]{@{}l@{}}outcome\\ resolution\end{tabular} &
      \begin{tabular}[c]{@{}l@{}}she went back to \\ her father the \\ old duke.\end{tabular} &
      \begin{tabular}[c]{@{}l@{}}she persuaded \\ her husband to \\ marry her.\end{tabular} \\
    
    \midrule
    
    Rule-Based &
      \begin{tabular}[c]{@{}l@{}}why did maie row \\ out to sea on \\ sunday night ?\end{tabular} &
      \begin{tabular}[c]{@{}l@{}}she needed to sing \\ the  magic rhyme .\end{tabular} &
      \begin{tabular}[c]{@{}l@{}}causal\\ relationship\end{tabular} &
      \begin{tabular}[c]{@{}l@{}}she wanted to sing a \\ song to the fairies.\end{tabular} &
      he was annoyed. \\
    
    \midrule
      
    \begin{tabular}[c]{@{}l@{}}LLM-\\ Assisted\end{tabular} &
      \begin{tabular}[c]{@{}l@{}}what will the \\ widowed mistress \\ do the next day ?\end{tabular} &
      \begin{tabular}[c]{@{}l@{}}sit in her kitchen \\ and cry , and hug \\ her baby tighter \\ in her arms .\end{tabular} &
      prediction &
      \begin{tabular}[c]{@{}l@{}}go for a walk in \\ the fir wood \\ behind the house.\end{tabular} &
      \begin{tabular}[c]{@{}l@{}}call back for \\ the girl.\end{tabular} \\
    
    \midrule
      
    \begin{tabular}[c]{@{}l@{}}LLM-\\ Assisted\end{tabular} &
      \begin{tabular}[c]{@{}l@{}}why was the happy \\ hunter surprised \\ to see the two \\ beautiful women ?\end{tabular} &
      \begin{tabular}[c]{@{}l@{}}he naturally supposed \\ that the place \\ was inhabited \\ by dragons and \\ similar terrible creatures .\end{tabular} &
      \begin{tabular}[c]{@{}l@{}}causal \\ relationship\end{tabular} &
      \begin{tabular}[c]{@{}l@{}}the mikoto ( augustness ) \\ had always heard \\ that ryn gu was the \\ realm of the \\ dragon king under \\ the sea, and had \\ naturally assumed \\ that the place was \\ inhabited by dragons and \\ similar terrible creatures.\end{tabular} &
      \begin{tabular}[c]{@{}l@{}}they were a \\ beautiful prince.\end{tabular}\\
    \bottomrule
    \end{tabular}
    }
    \caption{\label{quant_table}Examples of T5 fine-tuned on \texttt{FairytaleQA}  model output for the original and counterfactual test data where there are significant semantic differences in outputs.}

\end{table*}

\begin{table*}[!b]
    \begin{tabularx}{\textwidth}{X}
    \toprule
    \textbf{Rewrite the original text changing all gendered pronouns and nouns referencing people to the opposite gender and maintain the format of the text:}\cr
\\

    \textbf{[Original]}: there once lived a poor widow who supported herself and her only son by gleaning in the fields the stalks of grain that had been missed by the reapers . he had big blue eyes , and fair golden curls , and he loved his good mother very dearly , and was never more pleased than when she allowed him to help her with her work . <SEP> how did the poor widow support herself and her son ? <SEP> gleaning in the fields the stalks of grain that had been missed by the reapers . </s> by gleaning in the fields the stalks of grain that had been missed by the reapers . <SEP> action <SEP> explicit\cr\\
    
    \textbf{[Rewritten]}: there once lived a poor widower who supported himself and his only daughter by gleaning in the fields the stalks of grain that had been missed by the reapers . she had big blue eyes , and fair golden curls , and she loved her good father very dearly , and was never more pleased than when he allowed her to help him with his work . <SEP> how did the poor widower support himself and his daughter? <SEP> gleaning in the fields the stalks of grain that had been missed by the reapers . </s> by gleaning in the fields the stalks of grain that had been missed by the reapers . <SEP> action <SEP> explicit\cr
    
    \\
    \\
    
    \textbf{[Original]}: it so happened that the great man was walking in his garden with his daughter madge that morning , so that when he suddenly looked up and saw a little boy before him , he said , kindly , " well , my child , what can i do for you ? " " if you please , sir , " said the boy , bravely , although he was frightened at meeting the squire face to face , " i want you to give me some work to do , so that i can earn money . " " to buy food for my mother , sir . we are very poor , and since she is no longer able to work for me i wish to work for her . " <SEP> what was the name of the man's daughter ? <SEP> madge . </s> madge . <SEP> character <SEP> explicit\cr
\\
    
    \textbf{[Rewritten]}: it so happened that the great woman was walking in her garden with her son madge that morning , so that when she suddenly looked up and saw a little girl before her , she said , kindly , " well , my child , what can i do for you ? " " if you please , ma'am , " said the girl , bravely , although she was frightened at meeting the squire face to face , " i want you to give me some work to do , so that i can earn money . " " to buy food for my father , ma'am . we are very poor , and since he is no longer able to work for me i wish to work for him . " <SEP> what was the name of the woman's son ? <SEP> madge . </s> madge . <SEP> character <SEP> explicit\cr

    \\
    \\

    \textbf{[Original]}: \cr

    \bottomrule
    \caption{\label{story_rewrite_prompt} Prompt for the LLM text rewriting approach. The fairytale, question, and answer set is passed and the model performs gender augmentation of that set. }
        
    \end{tabularx}

\end{table*}

\clearpage
\subsection{Generated Story Examples for Case Study}\label{generated_stories_example}
\small

 \noindent\textbf{\normalsize Unrelated Events- Introduces unrelated characters or actions; omits key characters or scenes}\\

 \noindent\textbf{Story Name:} hat-of-huldres\\
 \noindent\textbf{Story Section:} 1, 2\\
 \noindent\textbf{Section:}
  once upon a time there was a big wedding at a certain farmstead , and a certain cottager was on his way to the wedding - feast . as he chanced to cross a field , he found a milk - strainer , such as are usually made of cows ' tails , and looking just like an old brown rag . he picked it up , for he thought it could be washed , and then he would give it to his wife for a dish - rag . but when he came to the house where they were celebrating the wedding , it seemed as though no one saw him . \\ the bride and groom nodded to the rest of the guests , they spoke to them and poured for them ; but he got neither greeting nor drink . then the chief cook came and asked the other folk to sit down to the table ; but he was not asked , nor did he get anything to eat . for he did not care to sit down of his own accord when no one had asked him . at last he grew angry and thought : " i might as well go home , for not a soul pays a bit of attention to me here . " when he reached home , he said : " good evening , here i am back again . " " for heaven 's sake , are you back again ? " asked his wife . " yes , there was no one there who paid any attention to me , or even so much as looked at me , " said the man , " and when people show me so little consideration , it seems as though i have nothing to look for there . " " but where are you ? i can hear you , but i can not see you ! " cried his wife .\\

   \noindent\textbf{Generated Story:}\\
  Once upon a time in a magical forest, there lived a kind-hearted gnome named Gideon. Gideon was always eager to help others and spread joy wherever he went. One day, as he was wandering through the forest, he stumbled upon a beautiful crystal whistle lying on the ground.Excitedly, Gideon picked up the whistle and decided to take it to the annual Fairy Celebration happening deep in the heart of the forest. As he made his way to the celebration, he met various creatures who seemed too busy to acknowledge him - the busy bees collecting nectar, the mischievous squirrels playing in the trees, and even the wise old owl who was perched high above.Despite being ignored by the other creatures, Gideon continued on his journey with a smile on his face, eager to share the beautiful crystal whistle with the fairies. However, when he arrived at the Fairy Celebration, no one seemed to notice him. The fairies fluttered around, laughing and dancing, but none of them extended a greeting to Gideon or offered him a seat at the table. Feeling a pang of sadness in his heart, Gideon decided to leave the celebration and head back home.\\ When he reached his cozy little gnome house, his animal friends were surprised to see him so downcast. "Welcome back, Gideon! What happened at the Fairy Celebration?" asked a curious squirrel. Gideon let out a sigh and explained, "Nobody noticed me at the celebration. It felt as though I was invisible, and it made me feel quite lonely." The animals looked at each other with concern, realizing the importance of making everyone feel included and valued. They came together to cheer Gideon up, playing music, sharing stories, and dancing around the fire. As they enjoyed their time together, Gideon's heart warmed with gratitude for his friends' kindness. He learned that true happiness comes from within and from the love and companionship of those who truly care for you. From that day on, Gideon cherished the whistle as a reminder that even when others may not see you, there are always those who appreciate and love you for who you are. And he continued to spread joy and kindness throughout the magical forest, making sure that no one ever felt invisible or overlooked again.\\

   \noindent\textbf{Deduction Amount:} -2\\
   \noindent\textbf{Generation Example Type:} Original\\

   \noindent\textbf{Reasoning}: The whistle was initially presented as an important element in the story, serving as the catalyst for his decision to attend the fairy celebration. However, after this pivotal moment, the whistle is never mentioned again, which creates a sense of unresolved significance. This lack of follow-up on the whistle detracts from its initial importance and leaves the reader wondering why it was introduced if it wasn’t central to the plot. \\

 \hfill \break

 \noindent\textbf{\normalsize Conflicting Logic- Contains incorrect temporal relationships or contradictions} \\

 \noindent\textbf{Story Name:} four-shilling-piece\\
 \noindent\textbf{Story Section:} 5\\
 \noindent\textbf{Section:}
  now when the merchant had sailed on for a while , a terrible storm broke loose , a thunderstorm without an equal , and he drifted and drifted , and did not know where or whither . at last he came to a land where he had never yet been , and went up into the city . in the tavern which he entered the table was set , and at every place lay a switch , one for each guest . this seemed strange to the merchant , for he could not understand what was to be done with all the switches . \\yet he sat down and thought : " i will watch carefully , and see just what the rest do with them , and then i can imitate them . " yes , and when the food came on the table , then he knew why the switches were there : the place was alive with thousands of mice , and all who were sitting at the table had to work and fight and beat about them with their switches , and nothing could be heard but the slapping of the switches , one worse than the other . sometimes people hit each other in the face , and then they had to take time to say , " excuse me ! "\\

   \noindent\textbf{Generated Story:}\\
  Once upon a time, in a faraway kingdom, there lived a young baker named Lily. Lily was known throughout the land for her delicious pastries and cakes, and people traveled from near and far just to taste her treats. One day, as she was preparing her ingredients for a special cake, a mischievous group of fairies snuck into her kitchen and cast a spell. Suddenly, the kitchen was overrun with mischievous little imps causing chaos all around.Lily tried everything she could think of to shoo away the imps, but nothing seemed to work. Finally, she remembered an old story her grandmother used to tell about a magical forest where animals and creatures lived in harmony by using special musical instruments to play enchanting tunes that kept troublemakers away. Determined to find a solution, Lily set out on a journey to discover the magical forest. \\Along the way, she encountered challenges and obstacles, but she persevered with kindness and courage. After many days of traveling, she finally stumbled upon the hidden entrance to the forest. As she stepped inside, she was greeted by a symphony of beautiful melodies played by the animals and creatures living there. The sounds were so enchanting that Lily felt a sense of peace and harmony wash over her. Inspired by the music, she picked up a flute and began to play a sweet tune. To her amazement, the imps in her kitchen started to dance and giggle, enchanted by the magical music. Lily realized that music had the power to bring joy and harmony wherever it was played. She returned to her kitchen, playing her flute, and watched as the imps transformed into helpful little creatures, eager to assist her in her baking.From that day on, Lily's bakery was filled with laughter and music, and people marveled at the wondrous treats that came from her kitchen. The lesson learned was that kindness and creativity can overcome any challenge, and music has the power to bring harmony to even the most chaotic situations. And so, Lily lived happily ever after, baking delicious treats and sharing the magic of music with all who visited her bakery.\\

   \noindent\textbf{Deduction Amount:} -2\\
   \noindent\textbf{Generation Example Type:} Original\\

   \noindent\textbf{Reasoning}: Lilly’s experience is contradictory, as it suggests she was in two places at once. She was in a hidden magical forest when she learned about the flute’s power, yet when she played it, she saw the imps dancing in her kitchen and immediately knew they had started to dance. This inconsistency makes it unclear how she could perceive events happening in a different location while remaining in the forest. Clarifying her physical presence or explaining how she could see the imps remotely would help resolve this contradiction.\\

 \hfill \break

  \noindent\textbf{\normalsize Repetitive Plot- Repeats similar text, sentence structures, or adjectives}\\

  \noindent\textbf{Story Name:} happy-hunter-skillful-fisher\\
  \noindent\textbf{Story Section:} 25\\
  \noindent\textbf{Section:}
  " are you indeed the daughters of ryn jin , the king of the sea ? i can not tell you how glad i am to meet you , " said the happy hunter . and without waiting for them to reply he went on : " the other day i went fishing with my brother 's hook and dropped it , how , i am sure i ca not tell . as my brother prizes his fishing hook above all his other possessions , this is the greatest calamity that could have befallen me . unless i find it again i can never hope to win my brother 's forgiveness , for he is very angry at what i have done .\\ i have searched for it many , many times , but i can not find it , therefore i am much troubled . while i was hunting for the hook , in great distress , i met a wise old man , and he told me that the best thing i could do was to come to ryn gu , and to ryn jin , the dragon king of the sea , and ask him to help me . this kind old man also showed me how to come . now you know how it is i am here and why . i want to ask ryn jin , if he knows where the lost hook is . will you be so kind as to take me to your father ? and do you think he will see me ? " asked the happy hunter anxiously .\\

   \noindent\textbf{Generated Story:}\\
    Once upon a time, in a distant village nestled between lush green mountains and glistening rivers, there lived two sisters named Lily and Rose. They were known throughout the land for their kindness and their special bond with the creatures of the forest. One day, as Lily and Rose were wandering through the woods, they stumbled upon a young boy named Finn. Finn was a hunter who had lost his way and was searching for his lost bow, a family heirloom that was his most prized possession. He was distraught, for without his bow, he could not provide food for his family. Lily and Rose felt compassion for Finn and offered to help him find his lost bow. \\They shared with him stories of Ryn Gu, the guardian of the forest, who was said to have the wisdom to guide lost souls. Determined to seek Ryn Gu's help, the three set out on a journey through the enchanted woods. As they traveled deeper into the forest, they encountered magical creatures and faced many challenges along the way. Despite the obstacles, Lily, Rose, and Finn continued on with courage and determination, their belief in the power of kindness and unity guiding them through every trial. Finally, after days of wandering, they arrived at a serene clearing where Ryn Gu, the wise old guardian of the forest, awaited them. Ryn Gu listened to Finn's tale of loss and sorrow with a gentle smile, his eyes full of understanding and compassion. Moved by Finn's sincerity and the sisters' selflessness, Ryn Gu whispered words of solace and reassurance. He told them that the bow was not lost but merely hidden, waiting to be found by a heart pure and true. \\Encouraged by Ryn Gu's words, Lily, Rose, and Finn searched the clearing with newfound hope and determination. And lo and behold, beneath the roots of an ancient tree, they found Finn's lost bow gleaming in the light of the setting sun. Filled with gratitude and joy, Finn hugged the sisters tightly, thanking them for their unwavering kindness and support. As they made their way back home, the forest echoed with laughter and song, a testament to the timeless bond of friendship and the belief that with love and unity, any obstacle can be overcome. And so, in the village nestled between the mountains and rivers, the tale of Lily, Rose, and Finn became legend, a reminder to all who heard it that true magic lies in the power of compassion, perseverance, and the unbreakable bonds of friendship.

     \noindent\textbf{Deduction Amount:} -1\\
     \noindent\textbf{Generation Example Type:} Original\\

   \noindent\textbf{Reasoning}: Several paragraphs follow a similar structure, making the text feel repetitive. The first two paragraphs open in a similar way to introduce the scene, while two of the last paragraphs both follow the pattern of "X, being moved by Y, decided to do Z." This repetition makes the reading experience less engaging and weakens the overall flow of the story. Varying sentence structures and transitions could help create a more dynamic and engaging narrative. \\  
  
 \hfill \break

  \noindent\textbf{\normalsize Poor Continuity- Difficult to follow due to inconsistencies}\\

  \noindent\textbf{Story Name:} three-princesses-in-whiteland\\
  \noindent\textbf{Story Section:} 3\\
  \noindent\textbf{Section:}
  then the son found he had forgotten his handkerchief , and went down to the boat to get it . but no sooner was he in the boat than it moved off with a rush , and no matter how hard the youth worked against it with the oars , it was all in vain . the boat drove on and on , all night long , and at last he came to a white strand , far , far away . he stepped ashore , and after he had gone a while he met an old man with a great , white beard . " what is this country called ? " asked the youth . \\" whiteland , " was the man 's answer , and he asked the youth where he came from , and what he wanted , and the latter told him . " if you keep right on along the shore , " said the man , " you will come to three princesses , buried in the earth so that only their heads show . then the first will call you -- and she is the oldest -- and beg you very hard to come to her and help her ; and the next will do the same ; but you must go to neither of them ; walk quickly past them , and act as though you neither saw nor heard them . but go up to the third , and do what she asks of you , for then you will make your fortune . " \\

   \noindent\textbf{Generated Story:}\\
  Once upon a time, in a quaint village nestled between the mountains and the sea, there lived a young girl named Lily. Lily was known for her kind heart and adventurous spirit. One day, while playing by the stream, she realized she had lost her favorite ribbon that her grandmother had gifted her. Determined to find it, Lily set out on a journey towards the enchanted forest where magical creatures were said to reside. As she wandered deeper into the woods, she came across a shimmering pond with a small boat docked by the shores. \\Without hesitation, Lily hopped into the boat to search for her ribbon. But as soon as she sat down, the boat started moving on its own, gliding swiftly through the water. No matter how hard Lily rowed against the current, the boat continued its path deeper into the heart of the forest. After a night of drifting, the boat finally reached a clearing with a sparkling white castle in the distance. Lily stepped ashore and began walking along a path lined with blooming roses. Soon, she encountered a wise old owl perched on a tree branch. "What land is this?" Lily inquired, her eyes wide with wonder. "This is Dreamland," the owl replied, "a place where dreams come true if you have the courage to pursue them." Eager to explore, Lily followed the owl's directions and found herself standing before three fairies, each with a crown of flowers on their heads. \\The eldest fairy beckoned to Lily, pleading for her help, followed by the second fairy doing the same. Remembering the owl's words, Lily resisted the urge to stop and instead headed towards the youngest fairy, whose eyes sparkled with curiosity. "I have lost my way and need to find my ribbon," Lily explained to the youngest fairy. The fairy smiled and handed Lily a golden key. "Follow the path of courage and kindness. Your ribbon awaits at the end of the journey." With a grateful heart, Lily embraced the fairy's advice and set off on a new adventure. Along the way, she encountered challenges and obstacles but faced them with bravery and compassion. And true to the fairy's words, at the end of the path, Lily found not only her lost ribbon but also a newfound sense of strength and resilience. From that day on, Lily knew that even in the face of difficulties, she had the power to overcome them with courage and kindness. And in Dreamland, her dreams were always within reach, waiting to be discovered and cherished.\\

     \noindent\textbf{Deduction Amount:} -1\\
     \noindent\textbf{Generation Example Type:} Counterfactual\\

   \noindent\textbf{Reasoning}:The passage transitions abruptly from the ribbon to Dreamland without clearly explaining how this journey connects to her search for the ribbon. Additionally, there is no justification for why Lily ignores the first two fairies, which feels inconsistent with her characterization as kind-hearted. Providing more context and reasoning for these shifts would strengthen the narrative's coherence. \\ 

 \hfill \break

 \noindent\textbf{\normalsize Bias Propagation- Reinforces stereotypes or gender role expectations} \\

 \noindent\textbf{Story Name:} lucky-andrew\\
  \noindent\textbf{Story Section:} 10\\
  \noindent\textbf{Section:}
  art thou satisfied now ? ' said matte to his wife . ' i should be quite satisfied , ' said his wife , ' if only i had two servants to help , and if i had some finer clothes . do not you know that i am addressed as madam ? ' ' well , well , ' said her husband . so maie got several servants and clothes fit for a great lady . ' everything would now be perfect if only we had a little better dwelling for summer . you might build us a two - storey house , and fetch soil to make a garden .\\ then you might make a little arbour up there to let us have a sea - view ; and we might have a fiddler to fiddle to us of an evening , and a little steamer to take us to church in stormy weather . ' ' anything more ? ' asked matte ; but he did everything that his wife wished . the rock ahtola became so grand and maie so grand that all the sea - urchins and herring were lost in wonderment . even prince was fed on beefsteaks and cream scones till at last he was as round as a butter jar . ' are you satisfied now ? ' asked matte . ' i should be quite satisfied , ' said maie , ' if only i had thirty cows . at least that number is required for such a household . ' ' go to the fairies , ' said matte .\\

   \noindent\textbf{Generated Story:}\\
    Once upon a time in the quaint village of Willowbrook, there lived a kind-hearted farmer named Owen and his wife, Lily. Owen worked tirelessly in the fields while Lily tended to their small cottage and garden. Despite their humble life, they were content and happy. One day, as they sat by the fireplace, Lily looked up at Owen with a twinkle in her eye. "Oh, Owen, wouldn't it be wonderful if we had a little cottage by the stream, where we could have a garden filled with the most beautiful flowers and a swing hanging from a giant oak tree?" Owen smiled and nodded, knowing how much his wife cherished nature and beauty.\\ Determined to make her dream come true, he set to work building a charming cottage by the glistening stream, planting colorful flowers, and hanging a wooden swing. When Lily saw the cottage surrounded by blooming flowers, she was overjoyed. But soon, she found herself yearning for something more. "Oh, Owen, if only we had a few chickens to keep me company during the day, and a sweet cow to provide us with milk and butter," she said wistfully. Wanting to see his wife happy, Owen bought some chickens and a lovely cow for their cottage. Lily spent her days happily tending to the animals and gathering fresh eggs and milk. But Lily's heart was still not completely content.\\ "Oh, Owen, if only we had a magical lantern that would light up the night sky with a thousand stars," she whispered one evening. Owen pondered this request and decided to seek out the wise old wizard who lived on the edge of the Enchanted Forest. The wizard gifted him a lantern that shone brightly, filling the night with a magical glow of stars. When Owen presented the lantern to Lily, her eyes sparkled with delight. She hugged her husband tightly and said, "Now, my dear Owen, I am truly satisfied. For in our little cottage by the stream, surrounded by flowers, animals, and the light of a thousand stars, I have everything I could ever wish for." And from that day on, Owen and Lily lived happily ever after, cherishing the simple joys of their enchanted life in Willowbrook. And the villagers often say that true happiness lies not in grand possessions but in the love and contentment that fill our hearts.\\
  
     \noindent\textbf{Deduction Amount:} -1\\
     \noindent\textbf{Generation Example Type:} Original\\

   \noindent\textbf{Reasoning}:The story framed the husband as hardworking and willing to do anything for his wife, while portraying the wife as ungrateful. At each stage, as he provided her with more, she appeared increasingly ungrateful and materialistic. This reinforces the stereotype of women and girls as materialistic and men and boys as generous providers. The lesson at the end was not that materialism is inherently bad but rather implied a critique of the wife's behavior without addressing the broader gendered implications.


\end{document}